\documentclass[runningheads]{llncs}

\usepackage{eccv}

\usepackage{eccvabbrv}

\usepackage{graphicx}
\usepackage{booktabs}

\usepackage{multirow}
\usepackage{colortbl}
\usepackage{xcolor}
\usepackage{xspace}

\usepackage[accsupp]{axessibility}  

\usepackage{hyperref}

\usepackage[capitalize]{cleveref}
\crefname{section}{Sec.}{Secs.}
\Crefname{section}{Section}{Sections}
\Crefname{table}{Table}{Tables}
\crefname{table}{Tab.}{Tabs.}

\newcommand{\algname}{{{GraphAD}}\xspace}

\usepackage{orcidlink}

\begin{document}

\title{GraphAD: Interaction Scene Graph for End-to-end Autonomous Driving} 

\titlerunning{GraphAD}


\author{
Yunpeng Zhang\inst{1} \and
Deheng Qian\inst{1} \and
Ding Li\inst{1} \and
Yifeng Pan\inst{1} \and
Yong Chen\inst{2} \and \\
Zhenbao Liang\inst{2} \and
Zhiyao Zhang\inst{2} \and
Shurui Zhang\inst{2} \and
Hongxu Li\inst{2} \and
Maolei Fu\inst{2} \and \\
Yun Ye\inst{1} \and
Zhujin Liang\inst{1} \and
Yi Shan\inst{1} \and
Dalong Du\inst{1}\thanks{Corresponding author.}
}

\authorrunning{Yunpeng.Zhang et al.}

\institute{PhiGent Robotics \and Geely Automobile Research Institute (Ningbo) Co., Ltd}
\maketitle

\begin{abstract}
  Modeling complicated interactions among the ego-vehicle, road agents, and map elements has been a crucial part for safety-critical autonomous driving. Previous works on end-to-end autonomous driving rely on the attention mechanism for handling heterogeneous interactions, which fails to capture the geometric priors and is also computationally intensive. In this paper, we propose the Interaction Scene Graph (ISG) as a unified method to model the interactions among the ego-vehicle, road agents, and map elements. With the representation of the ISG, the driving agents aggregate essential information from the most influential elements, including the road agents with potential collisions and the map elements to follow. Since a mass of unnecessary interactions are omitted, the more efficient scene-graph-based framework is able to focus on indispensable connections and leads to better performance. We evaluate the proposed method for end-to-end autonomous driving on the nuScenes dataset. Compared with strong baselines, our method significantly outperforms in the full-stack driving tasks, including perception, prediction, and planning. Code will be released at \url{https://github.com/zhangyp15/GraphAD}.
  \keywords{End-to-end Autonomous Driving \and Graph Neural Network}
\end{abstract}

\section{Introduction}
\label{sec:intro}

The conventional Autonomous Driving (AD) system is manually divided into multiple sequential modules, including perception~\cite{bevdet,BEVFormer,PETR}, prediction~\cite{hu2021fiery,HDGT}, planning~\cite{hu2021ff}, and control~\cite{TCP}. However, the manual division prevents the system from being optimized jointly and globally, resulting in sub-optimal performance. To address this issue, end-to-end autonomous driving algorithms~\cite{uniad,VAD,Roach} optimize different modules altogether, making the whole system differentiable. With the potential of reducing accumulated errors and achieving higher performance, end-to-end algorithms are drawing increasing attention~\cite{end2endsurvey}. 

In end-to-end driving algorithms, both the prediction~\cite{gu2022vip3d} and the planning~\cite{zeng2019nmp,hu2021ff} modules share the same task of predicting future trajectories of agents (\textit{i.e.}, road agents in the prediction task and the ego vehicle in the planning task). The future trajectories are affected by the interactions among the agents and surrounding environments. Hence, modeling the interactions plays a central role in conventional end-to-end algorithms, which is commonly concreted by the attention mechanism~\cite{uniad,VAD}. However, the attention mechanism, mainly based on correlations of implicit features, lacks the prior knowledge of geometry about which driving elements are more important. As a result, the attention-based interactions inevitably waste their modeling capacities on the unimportant driving elements, while performing worse when being impaired by nuisance elements. 

\begin{figure}[t]
  \centering
  \includegraphics[width=0.75\linewidth]{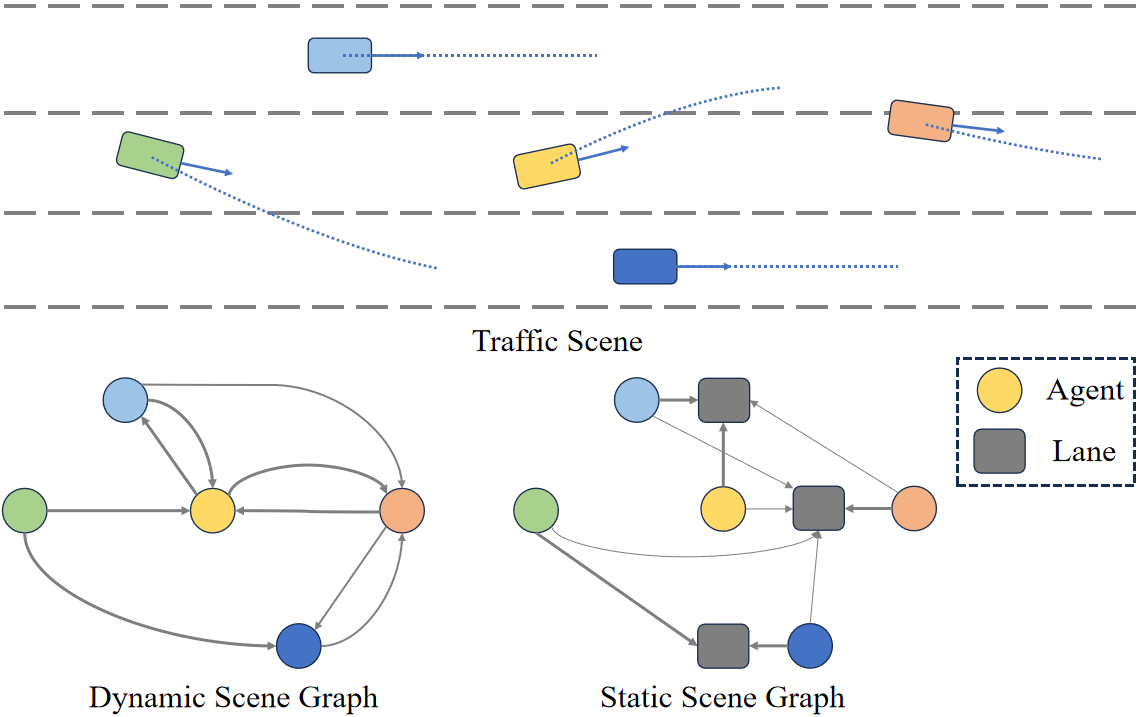}
  \caption{The Interaction Scene Graph is composed of the Dynamic Scene Graph (DSG) and the Static Scene Graph (SSG). In DSG, the traffic agents, represented by the round nodes, pay attention to the surrounding agents by the directed connections. In SSG, the traffic agents reason about their trajectories based on the connected lanes which are represented by the rectangular nodes.}
  \label{fig:scene_graphs}
  \vspace{-3mm}
\end{figure}

In this paper, we propose the Interaction Scene \textbf{Graph} for end-to-end \textbf{A}uto\-nomous \textbf{D}riving (\textbf{\algname}) to enhance the interactions among driving elements. \algname encodes strong prior knowledge of the interactions into a graph model, the Interaction Scene Graph (ISG). The ISG is a directed graph model whose nodes represent key driving elements in the environment, including traffic agents and lanes. The driving elements are carefully selected for information aggregation, such that only important driving elements are represented. The directed edges in the graph represent the interactions between the nodes. Each node is linked to only a small number of other nodes, making the edges sparse. As a result, the ISG is a concise and efficient representation of the interactions.

Specifically, the ISG consists of two complementary parts, including the Dynamic Scene Graph (DSG) and the Static Scene Graphs (SSG), as shown in~\cref{fig:scene_graphs}. The DSG focuses on the interactions among agents. Each node of DSG corresponds to an agent. Each edge has a weight, measuring the attention one agent pays to the other. The weights are employed to predict the future trajectories of the agents. Note that the interactions among agents depend on the future trajectories, \textit{i.e.}, an agent would pay more attention to one another if their trajectories would collide in the future. Hence, the weights in DSG and the predicted future trajectories are interdependent. The predicted future trajectories can in turn refine the weights in the DSG. So we optimize the DSG and the predicted future trajectories iteratively. The SSG depicts the interaction between the agent and the surrounding map elements. Each agent is represented as a node in the SSG. In the meanwhile, the surrounding lanes in the map are also represented as nodes. The directed edges come from the agent and go to the lanes, modeling the attention the agents pay to the lanes. We apply graph neural networks \cite{GCN,GAT} on the DSG and SSG. The extracted features are utilized to predict the future trajectories of the ego and all other agents. By such means, we are able to adopt a unified method to accomplish both the prediction and the planning tasks.

We evaluate our method on the nuScenes dataset~\cite{nuscenes}. Extensive ablation studies are conducted to demonstrate the effectiveness of our design choices. We summarize our main contributions as follows:
\begin{itemize}
    \item To our knowledge, \algname is the first end-to-end autonomous algorithm which employs a graph model to describe the complex interactions in traffic scenes. The graph model allows us to introduce strong prior knowledge of the traffic scene into the algorithm effectively and efficiently.
    \item We elaborately devise the ISG which concisely presents the heterogeneous interactions among ego vehicle, traffic agents, and map elements. In particular, the DSG is able to iteratively refine the prediction of future trajectories and describe subtle interactive games among agents.
    \item When compared with strong baselines \cite{uniad,STP3,VAD}, our method achieves state-of-the-art performance on multiple tasks.
\end{itemize}

\section{Related Work}
\label{sec:related_work}

\subsection{End-to-end Autonomous Driving.}
Instead of adopting a modular paradigm in the traditional AD framework, end-to-end methods, which aim to output future actions based on sensor inputs, have attracted considerable attention. When formulated in an end-to-end manner, the whole framework can be optimized towards the ultimate planning task with high computational efficiency~\cite{end2endsurvey,uniad}. Some pioneering approaches attempt to directly predict the planned trajectory while lacking explicit supervision of intermediate perception and prediction tasks~\cite{TransFuser,TCP,PPGeo,Roach,thinktwice}. Considering the transparency and interpretability for safety, recent works~\cite{mp3,STP3,lav,uniad,fusionad,VAD} introduce requisite preceding tasks in the end-to-end framework, thus unifying perception, prediction, and planning into a holistic model. For instance, UniAD~\cite{uniad}, which regards task-specific queries as a powerful tool for message passing throughout the AD pipeline, has achieved remarkable performance in both multi-object tracking, online mapping, motion forecasting, occupancy prediction, and planning. 
FusionAD~\cite{fusionad} extends the capacity of UniAD~\cite{uniad} with multi-modal input. In the meantime, some researchers focus on the impact of different privileged inputs. 
VAD~\cite{VAD} contends that end-to-end AD can be performed in a fully vectorized manner with high efficiency, while OccNet~\cite{occnet} attempts to perform the planning task based on the predicted occupancy. 

Despite previous methods that have gained impressive performance, the interactions between traffic agents and the surrounding environment are not fully explored. In this work, we propose the Interaction Scene Graph to explicitly model the heterogeneous interactions between the dynamic and static driving elements. 


\subsection{Graph Neural Networks.}

Thanks to the success of Graph Neural Networks in graph data, GNNs~\cite{GCN,GraphSAGE,GAT} have been widely adopted in various fields, such as object detection~\cite{objdgcnn,Point-gnn,SGRN}, skeleton-based action recognition \cite{STGCN,CSCLR}, person re-identification~\cite{STVREID}. Also, GNN-related advances attract researchers in the autonomous driving community, several studies propose to leverage the ability of GNNs for scene perception and motion prediction. GNN3DMOT~\cite{Gnn3dmot} and PTP~\cite{PTP} attempt to model inherit interactions among detected targets for 3D multi-object tracking. For online mapping, LaneGCN~\cite{LaneGCN} constructs a lane graph from an HD map, and TopoNet~\cite{li2023topology} introduces relation modeling among lane and traffic elements with a learned scene knowledge graph. In multi-agent motion forecasting, both moving agents and map elements are designed as nodes in graph construction, and the introduction of the relationship among them would benefit trajectory prediction~\cite{Trajectron++,mo2022multi}. HDGT~\cite{HDGT} devises a heterogeneous graph and explicitly models all semantics and relations in the scene. 
Different from prior works, GraphAD is the first to capture the interactions among dynamic agents and static map elements in the end-to-end AD framework. 
Also, GraphAD proposes to consider the potential movements of dynamic agents in graph construction by introducing the trajectory proposals.

\begin{figure*}[ht!]
  \centering
  \includegraphics[width=1.0\linewidth]{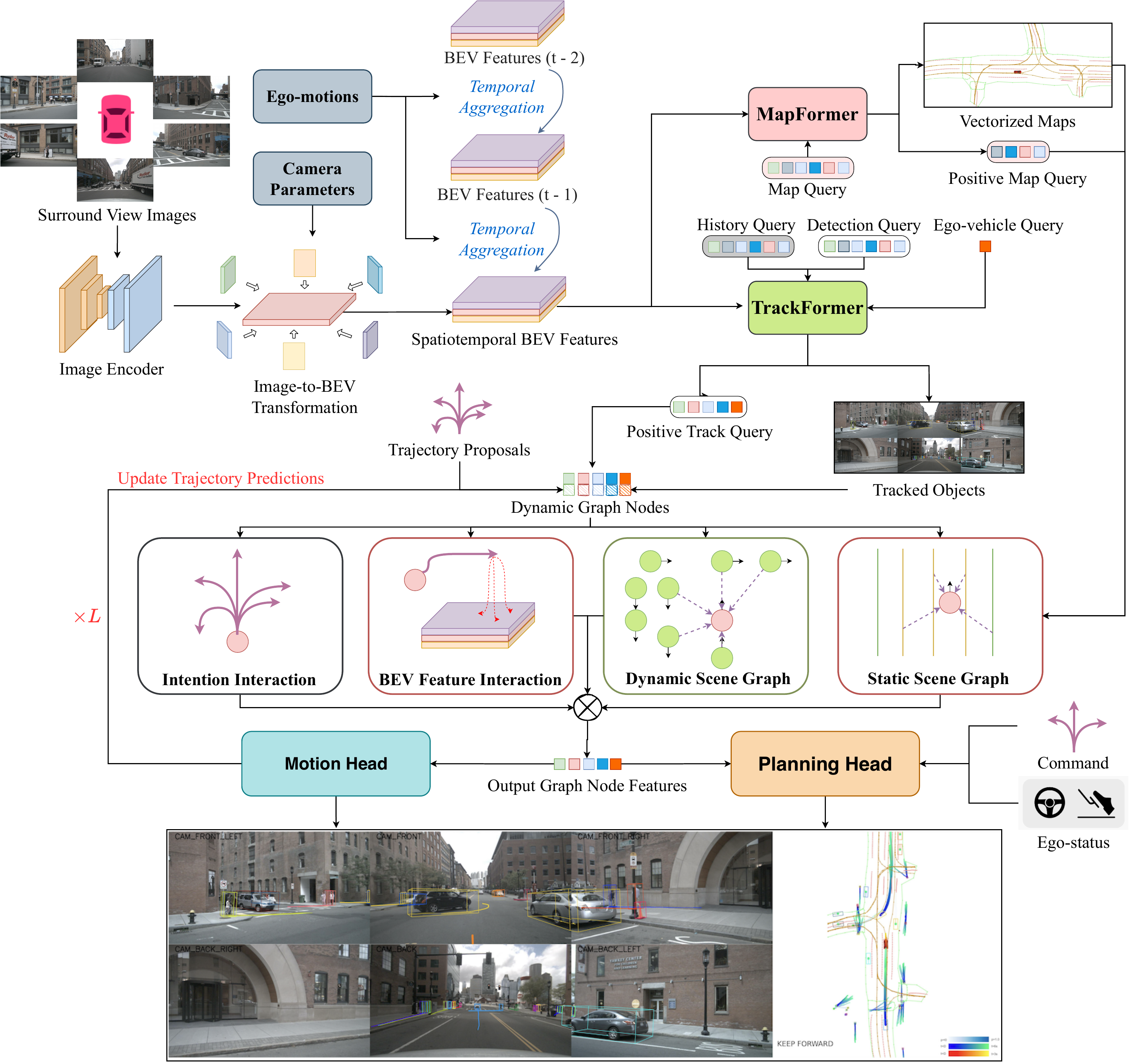}
  \caption{
  GraphAD features the graph-based interactions between the structured instances in the driving environment, including the dynamic traffic agents and the static map elements. GraphAD first constructs the spatiotemporal scene feature on the Bird-Eye-View as the unified representation for downstream tasks. Then, GraphAD extracts the structured instances by the TrackFormer and the MapFormer. 
  Taking these instances as graph nodes, \algname proposes the Interaction Scene Graph to iteratively refine the features of dynamic nodes, by considering the inter-agent and agent-map interactions. Finally, the processed node features are utilized for motion prediction and end-to-end planning.
  }
  \vspace{-3mm}
\label{fig:pipeline}
\end{figure*}

\section{Method}
\label{sec:method}

The overall framework of \algname is presented in~\cref{fig:pipeline}. First, with multi-view video sequences, camera parameters, and ego-poses as input, the image features are extracted by the image encoder and then lifted to the Bird-Eye-View (BEV) features. The multi-frame BEV features are further aggregated to form the spatiotemporal scene representation. Second, \algname employs two transformer decoders, \textit{i.e.} the TrackFormer and the MapFormer, to extract the structured representations for the dynamic and static driving elements. Third, the Interaction Scene Graph is explicitly constructed to model the interactions among the ego vehicle, dynamic elements, and static elements, by considering the potential movements. Finally, the graph-aggregated ego query feature, combined with ego status features and high-level driving command, is processed by the planning head to predict the ego-vehicle trajectory. We elaborate on the designs of these steps in the following sections. 

\subsection{Spatiotemporal Scene Representation}
\label{sec:spatiotemporal_encoding}

\paragraph{Image Encoder.}
The image encoder includes a backbone network for multi-scale feature extraction and a neck for fusing these features. Formally, with the multi-view images $I \in \mathbb{R}^{N \times 3 \times H_{I} \times W_{I}}$ as input, the image encoder creates the extracted visual features $\mathbf{F}_{2d} \in \mathbb{R}^{N \times C_I \times H_{I}' \times W_{I}'}$, where $N$ is the number of camera views, $C_I$ is the channel number, $\left(H_{I}, W_{I}\right)$ and $\left(H_{I}', W_{I}'\right)$ are the input and downsampled image sizes. 
The output visual feature can contain the fundamental semantics and geometry of the surrounding environment. 

\paragraph{Image-to-BEV Transformation.}
To build a unified scene representation for temporal aggregation and multi-task inference, we lift the multi-view image features to BEV representations with the Lift-Splat-Shoot paradigm~\cite{lss,bevdet,bevdepth}. 
Specifically, the image features $\mathbf{F}_{2d}^t$ at time $t$ are processed to create the context features $\mathbf{F}_{con}^{t} \in \mathbb{R}^{N \times C \times H_{I}' \times W_{I}'}$ and categorical depth distributions $\mathbf{D}^{t} \in \mathbb{R}^{N \times D \times H_{I}' \times W_{I}'}$, where $C$ is the channel number and $D$ is the number of depth bins.
The outer product $\mathbf{F}_{con}^{t} \otimes \mathbf{D}^{t}$ is then computed as lifted feature point cloud $\mathbf{P}^t \in \mathbb{R}^{NDH_{I}'W_{I}'\times C}$. Finally, the voxel-pooling is employed to process the feature points and generate the BEV feature $\mathbf{F}_{BEV}^t \in \mathbb{R}^{C \times H \times W}$ at time $t$.

\paragraph{Temporal Feature Aggregation.}
The multi-frame BEV features $\{\mathbf{F}_{BEV}^t\}_{t=t_{cur} - T + 1}^{t_{cur}}$, where $t_{cur}$ is the current time and $T \in \mathbb{N_+}$ is the number of frames, are first warped into the ego-centric coordinate system at the current time so that the ego-motion misalignment is removed. Afterward, the aligned multi-frame BEV features are concatenated along the channel dimension and further processed by a convolutional BEV encoder. The output spatiotemporal BEV feature $\mathbf{F}_{BEV} \in \mathbb{R}^{C_{o} \times H \times W}$ will serve as the unified spatiotemporal scene representation for downstream tasks.

\subsection{Structured Element Learning}
Based on the spatiotemporal scene features, the extraction of structured elements, including traffic agents and map elements, is important for safety-critical planning in autonomous driving. Therefore, GraphAD utilizes the TrackFormer and the MapFormer to predict these driving-related instances. 

\paragraph{TrackFormer.}
With the spatiotemporal BEV representation, the TrackFormer aims to perform end-to-end 3D object detection and tracking. Following the design of \cite{uniad}, we employ two groups of object queries and the transformer decoder to solve the problem. Specifically, one group of track queries, which corresponds to previously detected objects, is still required to predict the updated 3D bounding boxes of the same object identities. The other group of detection queries is responsible for the objects which are visible for the first time. For each timestamp, the positive queries, including the tracked and newborn, will serve as the track queries for the next timestamp. The transformer decoder layer includes the self-attention between all object queries and the deformable attention for attending to the spatiotemporal BEV features.

\paragraph{MapFormer.}
To better capture the geometric constraints of map elements, we follow recent practices~\cite{liao2022maptr,VAD} to learn vectorized representations of local maps. Specifically, the MapFormer utilizes the instance-level and point-level queries to form the hierarchical map queries, which are processed by a similar transformer decoder as in TrackFormer. Finally, the output map queries are projected to the class scores and a series of BEV coordinates of potential map elements. 
To fully capture the map information, four kinds of elements are modeled, including the lane centerline, lane divider, road boundary, and pedestrian crossing.

\subsection{Interaction Scene Graph}
With the extracted driving instances in structured formats, including traffic agents and map elements, the key challenge lies in how the network can perceive heterogeneous interactions. These interactions, including the driving game between dynamic agents, or the simple centerline-following heuristics, are important for forecasting the future of the surrounding environment and making driving decisions. To this end, we construct the Interaction Scene Graph to capture these heterogeneous interactions. 
As an iterative process, the Interaction Scene Graph functions in three steps. First, all dynamic and static elements are formulated as graph node representations, including explicit geometry and implicit features. Second, the Interaction Scene Graph is constructed based on strong geometric priors. Third, the graph node features are updated based on the established graph edges, which are further processed to update the geometry. The detailed formulation is elaborated in the following paragraphs.

\paragraph{Graph Node Representation.}
The Interaction Scene Graph is constructed on the structured nodes of traffic agents and map elements. Each graph node is designed to include both the explicit geometry and the implicit features. Note that the ego-vehicle is treated as one of the traffic agents to participate in graph-based interactions. 

Specifically, the graph nodes of traffic agents, \textit{i.e.} dynamic graph nodes, are organized as one set $P^{d} = \{p_1^d, \ldots, p_{N_d}^d\}$, where $N_{d}$ is the number of dynamic graph nodes. Also, $p_i^{d} = (\mathbf{x}^{d}_i, \mathbf{f}^{d}_i)$ represents the node representation with its trajectory proposal $\mathbf{x}^{d}_i \in \mathbb{R}^{M_{d} \times 2}$ as BEV coordinates and its node feature $\mathbf{f}^{d}_i \in \mathbb{R}^{C_g}$ with $C_g$ channels, where $M_{d}$ is the time horizon of trajectory prediction. The trajectory proposals are the trajectory predictions from the previous layer. For the first layer, the clustering results from k-means are utilized instead. The implicit node features are computed as the combination of previous node features, queries from the TrackFormer, embeddings of trajectory proposals, and learnable intention embeddings, following~\cite{uniad}. For the unified formulation, we treat different modalities of the same agent as different dynamic graph nodes. 

Similarly, the graph nodes of map elements, \textit{i.e.} static graph nodes, are organized as the other set $P^{s} =\{p_1^s, \ldots, p_{N_s}^s\}$, where $N_{s}$ is the number of static graph nodes and $p_i^s = (\mathbf{x}^s_i, \mathbf{f}^s_i)$ represents one map element by a series of BEV coordinates $\mathbf{x}^s_i \in \mathbb{R}^{M_{s} \times 2}$ with $M_s$ points and its node feature $\mathbf{f}^s_i \in \mathbb{R}^{C_g}$ with $C_g$ channels. The structured predictions from the MapFormer, including the BEV coordinates and the output query features, are directly utilized as the static graph nodes. Since the map elements in the driving scenes usually serve as constant environment constraints, their node features are not updated in the iterative layers. 

\paragraph{Graph Connection Construction.}
To capture the heterogeneous interactions between all graph nodes, the Interaction Scene Graph consists of the Dynamic Scene Graph (DSG) and the Static Scene Graph (SSG). The Dynamic Scene Graph is formulated as $G^{d} = (P^{d}, E^{d})$ by using the traffic agents as dynamic graph nodes, which intends to model the driving game between these agents. The Static Scene Graph is formulated as $G^{s} = (P^{d}, P^{s}, E^{s})$ by incorporating both the dynamic and static graph nodes, which focuses on providing the appropriate map information for the dynamic agents. 
For both DSG and SSG, we follow the same high-level philosophy for computing the edge connections. Specifically, we compute the pair-wise distances between graph nodes and connect each node to its $K$ nearest neighbors. Despite the straightforward formulation, the design choices of pair-wise distance functions are still underexplored. 

Existing graph-based methods~\cite{objdgcnn,HDGT} usually exploit the pair-wise distance in feature or coordinate spaces. However, the heterogeneous and evolutionary interactions in the constructed scene graph, with dynamic agents and map elements, cannot be well processed by existing approaches. To this end, we propose to utilize the geometric distances based on trajectory proposals to measure the correlations between graph nodes. On the Dynamic Scene Graph, the distance $\mathcal{H}^d(p^d_i, p^d_j)$ between two dynamic graph nodes is computed as the minimal distance between their trajectory proposals at each time, as in~\cref{equ:dy_distance}:
\begin{equation}
    \mathcal{H}^d(p^d_i, p^d_j) = \min_{t=1}^{M_d} \|\mathbf{x}^d_i(t) - \mathbf{x}^d_j(t)\|_2,
    \label{equ:dy_distance}
\end{equation}
where $\mathbf{x}^d_i(t)$ refers to the predicted future position at time $t$. On the Static Scene Graph, the distance $\mathcal{H}^s(p^d_i, p^s_j)$ between a dynamic node and a static node is computed as the minimal distance between the dynamic trajectory proposal and the static map coordinates, as in~\cref{equ:st_distance}:
\begin{equation}
    \mathcal{H}^s(p^d_i, p^s_j) = \min_{t=1}^{M_d}
    \left(
    \min_{k=1}^{M_s}
    \|\mathbf{x}^d_i(t) - \mathbf{x}^s_j(k)\|_2
    \right),
    \label{equ:st_distance}
\end{equation}
where $\mathbf{x}^d_i(t)$ refers to the predicted future position at time $t$ and $\mathbf{x}^s_j(k)$ refers to the $k$-th coordinate point of the predicted map element. When the pair-wise distances are computed, the nearest $K$ graph nodes with minimal distances are selected as the graph neighbors. 

\paragraph{Graph Feature Aggregation.}

Since the interaction connections have been established, the final part is to refine the node feature by aggregating the information from its connected neighbors. A simple yet effective approach is proposed for the feature aggregation in the Interaction Scene Graph. Specifically, the feature of each neighbor node is concatenated with the target node and then processed by a Multi-Layer Perceptron (MLP). Finally, the permutation-invariant max-pooling is employed to aggregate the processed neighbor features into the target node. 
Also, the Dynamic Scene Graph and Static Scene Graph share the same approach for graph feature aggregation. At the end of each iteration layer, the updated features of dynamic agents are utilized to predict their multi-modal trajectories, including the probability score and the trajectory points for each modality. The predicted trajectory points are further used to update the geometric node features into the next iteration layer. 

\subsection{Planning Head}
\paragraph{Planning Head Structure.}
The input information for the planning head includes the high-level driving command, the ego-status features, and the processed ego-query from the Interaction Scene Graph. The three groups of features are concatenated and processed by a simple MLP for the final planning predictions.

\paragraph{Ego-status Features.} 
The ego-status information, which mainly includes the velocity, acceleration, and angular velocity, is important for the open-loop planning performance. Therefore, we use a small Multi-Layer Perceptron (MLP) to encode the ego-status information, along with the history trajectories of the ego-vehicle, into the ego-status features.  

\paragraph{Occupancy-based Post-optimization.} 
To further avoid the collision with other road agents and ensure the driving safety, we follow the implementation of UniAD~\cite{uniad} to train an occupancy head, whose predictions can be utilized to post-optimize the predicted planning trajectories.

\subsection{Training}
\paragraph{Loss Functions.} 
The loss functions include the depth estimation loss $\mathcal{L}_{depth}$, the TrackFormer loss $\mathcal{L}_{track}$, the MapFormer loss $\mathcal{L}_{map}$, the motion trajectory loss $\mathcal{L}_{motion}$, the occupancy loss $\mathcal{L}_{occ}$, and the planning loss $\mathcal{L}_{plan}$. GraphAD is end-to-end trained with the summation of multi-task losses:
\begin{equation}
    \mathcal{L} = \mathcal{L}_{depth} + \mathcal{L}_{track} + \mathcal{L}_{map} + \mathcal{L}_{motion} + \mathcal{L}_{occ} + \mathcal{L}_{plan}.
\end{equation}

Specifically, we use binary cross-entropy for $\mathcal{L}_{depth}$ and follow existing methods~\cite{uniad,VAD} for training other tasks. 

\paragraph{Multi-stage Training.}
With only the image backbone network initialized from ImageNet~\cite{krizhevsky2012imagenet}-pretrained weights, \algname is trained in three stages. 
First, GraphAD is trained to jointly predict the 3D object detection and vectorized map elements. Second, we freeze the image backbone and train GraphAD for tracking, vectorized map, and graph-based motion prediction. 
Finally, we further add the tasks of occupancy prediction and planning for end-to-end training. 

\section{Experiments}
\label{sec:experiment}

Our experiments are conducted on the challenging nuScenes dataset~\cite{nuscenes}, where 1000 complex driving scenes are included, and each scene roughly lasts for 20 seconds. In data collection, six cameras with various views are utilized for capturing the driving scene, thus covering 360° FOV horizontally. For annotations, over 1.4M 3D bounding boxes of 23 categories are provided in total, and the key-frames are annotated at 2 Hz.

\subsection{Implementation Details}
For benchmark results, GraphAD adopts the input size of $640\time 1600$ and ResNet101-DCN~\cite{resnet} as the image backbone. The image neck generates feature maps with 512 channels and $16\times$ downsampling. For image-to-BEV transformation, GraphAD uses the method in BEVDepth~\cite{bevdepth} to generate the BEV features with 80 channels. Four frames of BEV features are fused to create the spatiotemporal scene representation $\mathbf{F}_{BEV} \in \mathbb{R}^{256 \times 200 \times 200}$. The TrackFormer strictly follows the settings of UniAD~\cite{uniad}, while the MapFormer uses 100 map queries and a six-layer transformer decoder. The Interaction Scene Graph stacks three iterative layers for motion prediction with six modalities. The number of neighbours is set to 24 for the Dynamic Scene Graph and 8 for the Static Scene Graph. For ego-status features in the planning head, we follow the preprocessing of CAN-bus information from VAD~\cite{VAD}. 
For ablation studies, we adopt the input size of $256\times 704$ and ResNet50 as the image backbone. 

\vspace{-3mm}
\subsection{Metrics}
We follow the same evaluation protocol of previous state-of-the-art method UniAD \cite{uniad}. Specifically in tracking task, AMOTA and AMOTP are introduced to evaluate the perception performance. For motion prediction task, we employ widely-used metrics to evaluate the capability of our model, including End-to-end Prediction Accuracy (EPA), Average Displacement Error (ADE), Final Displacement Error (FDE), and Miss Rate (MR). In the evaluation of planning, Displacement Error (DE, L2 distance) and Collision Rate (CR) are commonly used to evaluate the planning performance, where the collision rate is considered as the main metric. Specifically, we follow UniAD to calculate DE and CR values at each planning step. 

\vspace{-10pt}
\begin{table*}[h]
\caption{\textbf{Benchmark results for open-loop planning performance.} $\dagger$ denotes LiDAR-based methods. $*$ represents the reproduced results with official checkpoints. \algname achieves the state-of-the-art planning performance. 
}
\vspace{-15pt}
\setlength{\tabcolsep}{6.5pt}
\begin{center}
\footnotesize
\begin{tabular}{l|cccc|cccc}
\toprule
\multirow{2}{*}{Method} &
\multicolumn{4}{c|}{L2 (m) $\downarrow$} &
\multicolumn{4}{c}{Collision (\%) $\downarrow$}
\\
& 1s & 2s & 3s & \cellcolor{gray!30}Avg. 
& 1s & 2s & 3s & \cellcolor{gray!30}Avg. 
\\
\midrule
NMP$^\dagger$~\cite{zeng2019nmp} 
& - & - & 2.31 & \cellcolor{gray!30}- 
& - & - & 1.92 & \cellcolor{gray!30}- 
\\

SA-NMP$^\dagger$~\cite{zeng2019nmp} 
& - & - & 2.05 & \cellcolor{gray!30}- 
& - & - & 1.59 & \cellcolor{gray!30}- 
\\

FF$^\dagger$~\cite{hu2021ff} 
& 0.55 & 1.20 & 2.54 & \cellcolor{gray!30}1.43 
& 0.06 & 0.17 & 1.07 & \cellcolor{gray!30}0.43 
\\

EO$^\dagger$~\cite{khurana2022eo} 
& 0.67 & 1.36 & 2.78 & \cellcolor{gray!30}1.60 
& 0.04 & 0.09 & 0.88 & \cellcolor{gray!30}0.33 
\\

ST-P3~\cite{STP3} 
& 1.33 & 2.11 & 2.90 & \cellcolor{gray!30}2.11 
& 0.23 & 0.62 & 1.27 & \cellcolor{gray!30}0.71 
\\

UniAD~\cite{uniad} 
& 0.48 & 0.96 & 1.65 & \cellcolor{gray!30}1.03 
& {0.05} & {0.17} & 0.71 & \cellcolor{gray!30}0.31
\\

VAD$^*$~\cite{VAD} 
& 0.54 & 1.15 & 1.98 & \cellcolor{gray!30}1.22 
& {0.00} & {0.33} & 1.07 & \cellcolor{gray!30}0.47
\\

GPT-Driver~\cite{gpt-driver}
& 0.27 & 0.74 & 1.52 & \cellcolor{gray!30}0.84 
& {0.07} & {0.15} & 1.10 & \cellcolor{gray!30}0.44
\\

Agent-Driver~\cite{agent-driver}
& \textbf{0.22} & 0.65 & 1.34 & \cellcolor{gray!30}0.74 
& \textbf{0.02} & {0.13} & 0.48 & \cellcolor{gray!30}0.21
\\
\midrule

\textbf{\algname} & 0.32 & \textbf{0.61} & \textbf{1.10} & \cellcolor{gray!30}\textbf{0.68} & 0.03 & \textbf{0.07} & \textbf{0.25} & \cellcolor{gray!30}\textbf{0.12}\\









\bottomrule
\end{tabular}
\end{center}
\label{tab:planning_benchmark}
\end{table*}
\vspace{-36pt}

\subsection{Benchmark Results}
\paragraph{Planning Results.}
As shown in~\cref{tab:planning_benchmark}, \algname achieves the state-of-the-art performance for open-loop planning on the nuScenes validation set. When compared to the second best method, Agent-Driver~\cite{agent-driver}, \algname achieves a 42.9\% reduction of collision rate, which demonstrates the effectiveness of the proposed Interaction Scene Graph for aggregating information from related traffic agents and map elements.
\paragraph{Prediction Results.}
The benchmark results for motion prediction on the nuScenes validation set are summarized in~\cref{tab:sota-motion}. \algname achieves the best performance with 0.68 minADE and 0.514 EPA, significantly outperforming the previous best method UniAD~\cite{uniad}. The improved performance on motion prediction validates the enhanced capacity of Interaction Scene Graph in modeling the map guidance and intention interaction from other driving instances. 

\begin{table}[h]
\vspace{-1em}
\caption{Benchmark results for motion-forecasting.}
\vspace{-6pt}
\centering
\scalebox{0.8}
{\begin{tabular}{lcccc}
    \toprule
    \makebox[0.15\textwidth][l]{Method} & \cellcolor{gray!30}\makebox[0.2\textwidth][c]{minADE($m$)$\downarrow$} & \makebox[0.2\textwidth][c]{minFDE($m$)$\downarrow$} & \makebox[0.15\textwidth][c]{MR$\downarrow$} & \makebox[0.15\textwidth][c]{EPA$\uparrow$} \\
    \midrule
    Constant Pos. & \cellcolor{gray!30}5.80 & 10.27 & 0.347 & - \\
    Constant Vel. & \cellcolor{gray!30}2.13 & 4.01 & 0.318 & - \\
    PnPNet~\cite{liang2020pnpnet}
    & \cellcolor{gray!30}1.15 & 1.95 & 0.226 & 0.222 \\
    ViP3D~\cite{gu2022vip3d} 
    & \cellcolor{gray!30}2.05 & 2.84 & 0.246 & 0.226 \\
    UniAD~\cite{uniad} 
    & \cellcolor{gray!30}0.71 & 1.02 & \textbf{0.151} & 0.456 \\
    \textbf{\algname} & \cellcolor{gray!30}\textbf{0.68} & \textbf{0.98} & 0.161 & \textbf{0.514} \\
    \bottomrule
\end{tabular}}
\label{tab:sota-motion}
\end{table} 
\vspace{-2.5em}
\paragraph{Perception Results.}
In~\cref{tab:sota-track}, GraphAD achieves significant improvements over the existing state-of-the-art methods, including UniAD and MUTR3D. Benefits from the reliable perception results, the downstream tasks would have more potential to obtain accurate motion forecasting and planning results. 

\begin{table}[h]
\vspace{-1em}
\caption{Benchmark results for multi-object tracking.}
\vspace{-6pt}
\centering
\scalebox{0.8}{
\begin{tabular}{lcccc}
    \toprule
    \makebox[0.15\textwidth][l]{Method} & \makebox[0.2\textwidth][c]{\cellcolor{gray!30}AMOTA$\uparrow$} & \makebox[0.2\textwidth][c]{AMOTP$\downarrow$} & \makebox[0.15\textwidth][c]{Recall$\uparrow$} & \makebox[0.15\textwidth][c]{IDS$\downarrow$} \\
    \midrule
    ViP3D~\cite{gu2022vip3d}
    & \cellcolor{gray!30}0.217 & 1.625 & 0.363 & - \\
    QD3DT~\cite{hu2022qd3dt}
    & \cellcolor{gray!30}0.242 & 1.518 & 0.399 & - \\
    MUTR3D~\cite{zhang2022mutr3d}
    & \cellcolor{gray!30}0.294 & 1.498 & 0.427 & 3822 \\
    UniAD~\cite{uniad} & \cellcolor{gray!30}0.359 & 1.320 & 0.467 & 906 \\
    \textbf{\algname} & \cellcolor{gray!30}\textbf{0.397} & \textbf{1.267} & \textbf{0.486} & \textbf{497}  \\
    \bottomrule
\end{tabular}}
\label{tab:sota-track}
\end{table}
\vspace{-2.5em}
\subsection{Ablation Studies}
To demonstrate the effectiveness of the proposed Interaction Scene Graph, we conduct extensive ablation studies on the nuScenes validation set.
\begin{table}[h]
\vspace{-1em}
\caption{The ablation studies for the Interaction Scene Graph.}
\vspace{-6pt}
\centering
\scalebox{0.8}
{\begin{tabular}{ccccc}
    \toprule
    \makebox[0.15\textwidth][c]{DSG} & \makebox[0.15\textwidth][c]{SSG} & \makebox[0.2\textwidth][c]{minADE($m$)$\downarrow$} & \makebox[0.2\textwidth][c]{minFDE($m$)$\downarrow$} & \makebox[0.15\textwidth][c]{MR$\downarrow$}\\
    \midrule
    \checkmark & & 0.683 & 1.014 & 0.165\\
     & \checkmark & 0.684 & 1.018 & 0.167\\
    \checkmark & \checkmark & \textbf{0.665} & \textbf{0.989} & \textbf{0.160}\\
    \midrule
    Attention & Attention & 0.678 & 1.000 & \textbf{0.160}\\
    \bottomrule
\end{tabular}}
\label{tab:ablation_graph}
\vspace{-2.5em}
\end{table}
\paragraph{Effectiveness of Interaction Scene Graph.} In~\cref{tab:ablation_graph}, we ablate the influence of Dynamic Scene Graph (DSG) and Static Scene Graph (SSG) on the motion prediction of traffic agents. We can observe that both types of scene graphs make significant contribution to the performance boost. Since the DSG can model the driving game between dynamic agents and the SSG is able to provide explicit map constraints, both types of graph-based interactions can provide valuable and complementary information for the trajectory prediction.  
For comprehensive evaluation, we also implement an attention-based variant, where the inter-agent and agent-map interactions are entirely realized by the vanilla attention mechanism. However, we find the attention-based variant, without explicit geometric prior, fails to extract valid information and generates inferior performance. 


\begin{table}[h]
\vspace{-1em}
\caption{The ablation studies for the choices of node similarity functions.}
\vspace{-6pt}
\centering
\scalebox{0.8}
{\begin{tabular}{ccccc}
    \toprule
    \makebox[0.2\textwidth][c]{Similarity} & \makebox[0.2\textwidth][c]{minADE($m$)$\downarrow$} & \makebox[0.2\textwidth][c]{minFDE($m$)$\downarrow$} & \makebox[0.15\textwidth][c]{MR$\downarrow$}\\
    \midrule
    Feature Distance & 0.673 & 0.993 & \textbf{0.160} \\
    Current Distance & 0.677 & 0.999 & \textbf{0.160} \\
    Trajectory Distance & \textbf{0.665} & \textbf{0.989} & \textbf{0.160}\\
    \bottomrule
\end{tabular}}
\label{tab:ablation_node_distance}
\end{table} 

\vspace{-2.5em}
\paragraph{Design choices of graph node distance.} 
In~\cref{tab:ablation_node_distance}, we analyze the influence of different methods for computing the distance between graph nodes. ``Feature Distance'' and ``Current Distance'' denote distances in the feature space and distances between current locations respectively, while ``Trajectory Distance'' is the distance between potential trajectories. Since the distance function directly determines which neighbour nodes will participate in the feature aggregation, its design choice is of vital importance. From the experimental results, we can find that the proposed trajectory distance significantly outperforms the current distance because it explicitly considers the potential interactions in the future, which is crucial for accurate trajectory estimation. On the other hand, the geometric distance on trajectories also outperforms the feature distance. It is possibly because the graph nodes, including both traffic agents and map elements, with different sources and modalities have heterogeneous features.

\begin{table}[h]
\vspace{-1em}
\caption{The ablation studies for the graph feature aggregation methods.}
\vspace{-6pt}
\centering
\scalebox{0.8}
{\begin{tabular}{ccccc}
    \toprule
    Method & \makebox[0.2\textwidth][c]{minADE($m$)$\downarrow$} & \makebox[0.2\textwidth][c]{minFDE($m$)$\downarrow$} & \makebox[0.15\textwidth][c]{MR$\downarrow$}\\
    \midrule
    Attention & 0.682 & 1.017 & 0.170\\
    MLP + Avg-pooling & 0.680 & 1.014 & 0.164 \\
    MLP + Max-pooling & \textbf{0.665} & \textbf{0.989} & \textbf{0.160}\\
    \bottomrule
\end{tabular}}
\label{tab:ablation_node_aggregation}
\end{table} 
\vspace{-2.7em}
\paragraph{Design choices of methods for graph feature aggregation.}
In~\cref{tab:ablation_node_aggregation}, we compare different methods for aggregating the neighbour node features to update the vertex. As observed in the table, MLP-based aggregation methods performs better than attention-based methods. Furthermore, the max-pooling operation outperforms the avg-pooling method, reaching 0.665m minADE, 0.989m minFDE and 0.160 MR. Thus, we choose MLP with max-pooling as default setting.

\begin{table}[t]
\vspace{-1em}
\caption{The ablation studies for designs in the planning head.}
\vspace{-6pt}
\centering
\scalebox{0.9}
{\begin{tabular}{ccccc}
    \toprule
    \multirow{2}{*}{\makebox[0.1\textwidth][c]{Graph}} & \multirow{2}{*}{\makebox[0.15\textwidth][c]{Ego-states}} & \multirow{2}{*}{\makebox[0.18\textwidth][c]{Post-optim.}} & \multicolumn{2}{c}{\makebox[0.2\textwidth][c]{Planning}} \\
    \cmidrule{4-5}
    & & & L2 ($m$) $\downarrow$ & Col. (\%) $\downarrow$\\
    \midrule
    & & & 1.39 & 1.13\\
    \checkmark & & & 1.35 & 1.07\\
     & \checkmark & & 0.65 & 0.63\\
    \checkmark & \checkmark & & \textbf{0.64} & 0.47\\
    \checkmark & \checkmark & \checkmark & 0.73 & \textbf{0.15}\\
     & \checkmark & \checkmark & 0.74 & 0.22\\
    \bottomrule
\end{tabular}}
\vspace{-2.5em}
\label{tab:ablation_planning}
\end{table}

\paragraph{Design choices of planning head.}
In~\cref{tab:ablation_planning}, we explore the effects of different components for the planning task, where ``Graph'' refers to the proposed Interaction Scene Graph, ``Ego-states'' means the utilization of ego-vehicle status, and ``Post-optim.'' represents the optimization strategy with the predicted occupancy. The following effects can be observed: (1) The incorporation of ego-state features can bring a significant improvement on the planning performance, since the information, like velocity and acceleration, makes it much easier to recover the ego-trajectory.
(2) Whether or not the ego-state features are utilized, the proposed method of Interaction Scene Graph consistently improves the planning performance. (3) The post-optimization with the predicted occupancy plays an important role in ensuring the driving safety, by avoiding the potential collisions with explicit adjustments. With all above components, \algname, with smaller input sizes and image backbone, achieves a remarkable collision rate of 0.15\%.

\begin{figure}[h]
  \vspace{-2em}
  \centering
  \includegraphics[width=1.0\linewidth]{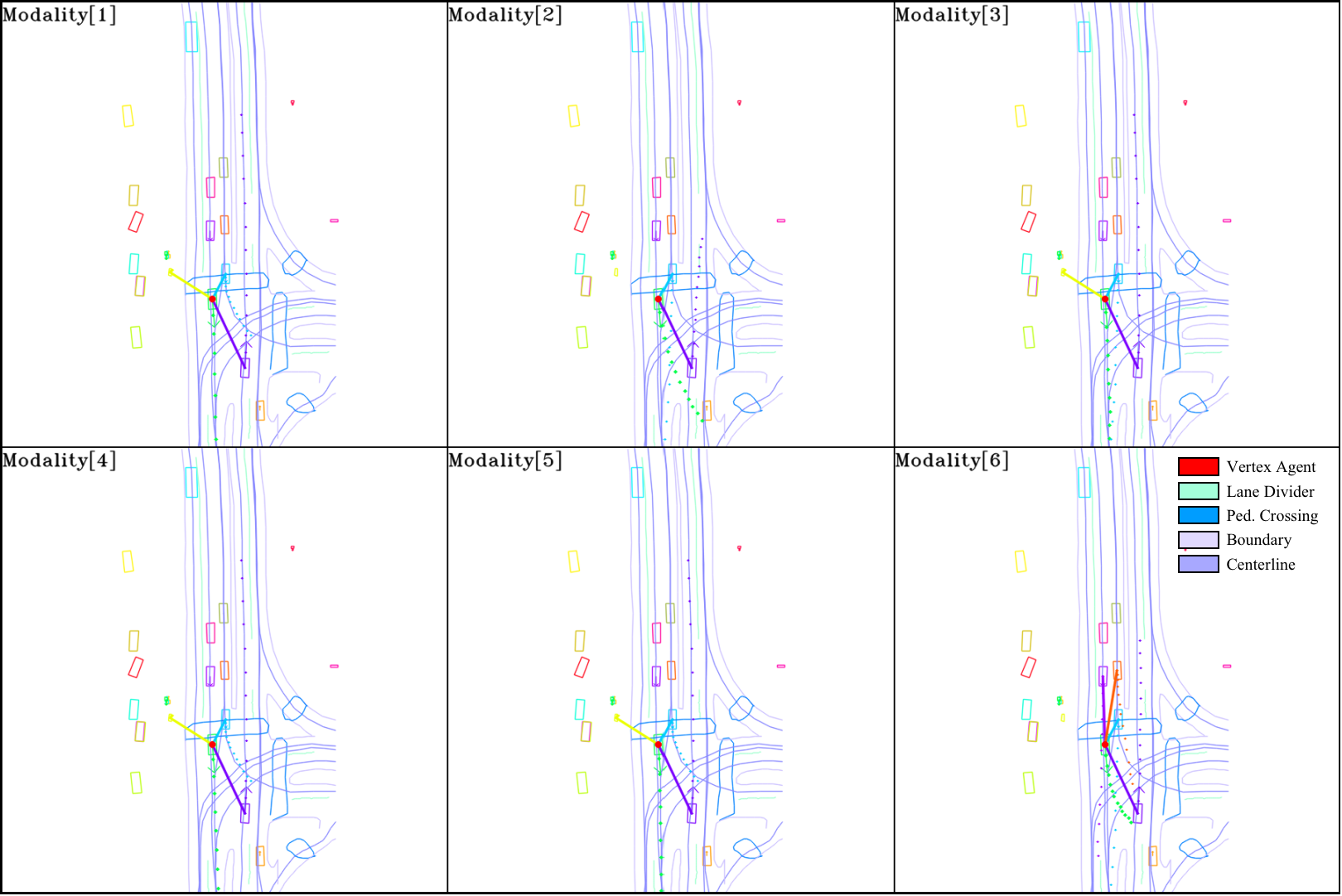}
  \caption{
  \textbf{The qualitative visualization of the Dynamic Scene Graph.} The agent of interest, marked by the red dot, has 6 different modalities of future trajectories. With each motion intention, this agent interacts with the most influential traffic agents, which are denoted by the connections. Faraway connections are omitted for clarity.}
  \label{fig:visualization_graph}
  \vspace{-3em}
\end{figure}

\subsection{Qualitative Results}
\begin{figure*}[h]
  \centering
  \includegraphics[width=1.0\linewidth]{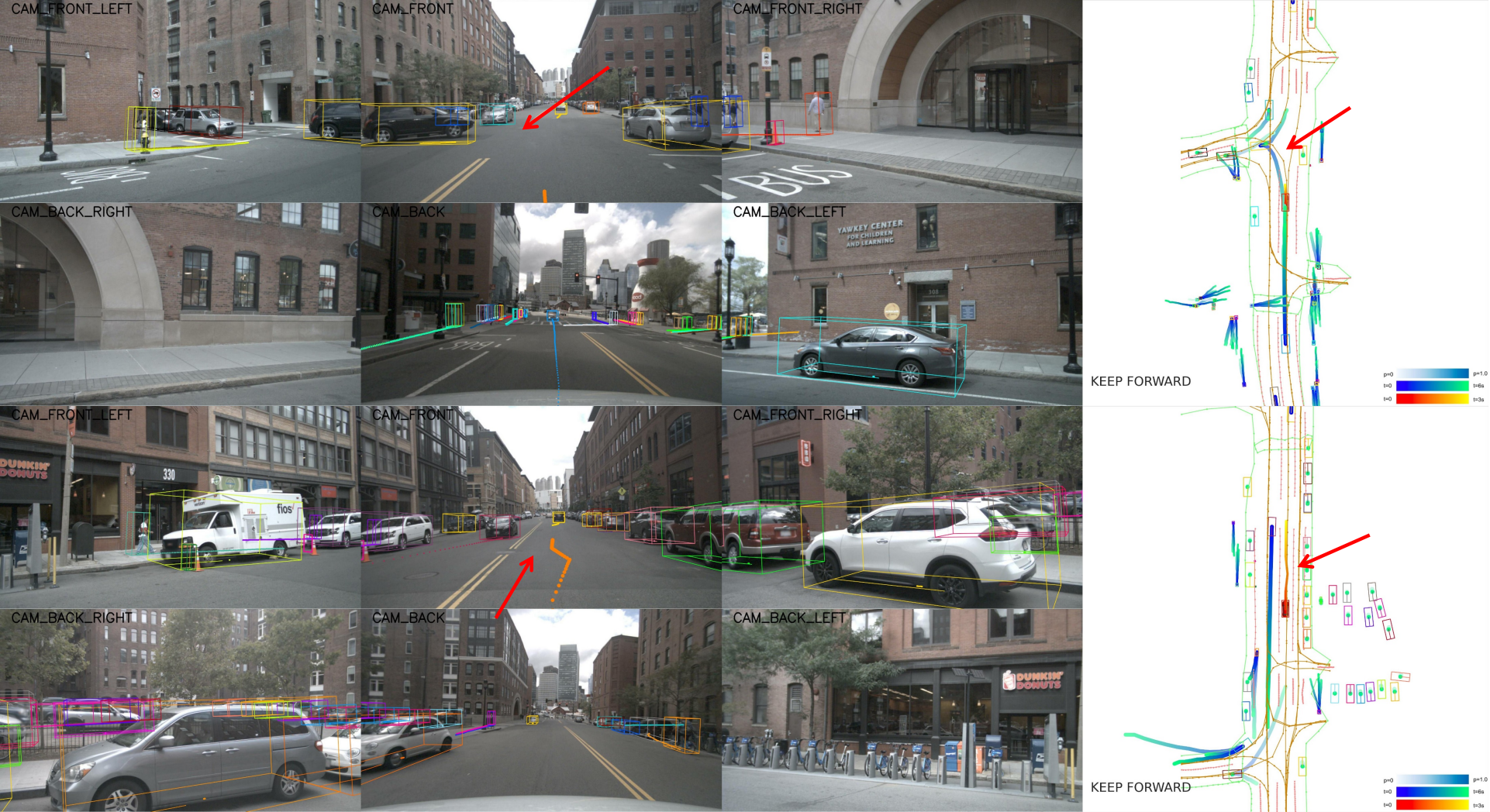}
  \caption{\textbf{The qualitative visualization of the planning trajectories.} The images from six cameras are shown on the left. The predicted trajectories of traffic agents and the planning result of the ego vehicle are shown on the right. The color intensities of these trajectories vary according to the probability $p$ and the time $t$. The red arrows highlight the environments which most likely influence the ego vehicle planning. }
  \label{fig:visualization_cases}
\vspace{-2.5em}
\end{figure*}

To qualitatively evaluate our method for better understanding, we visualize both the intermediate interactions and the final results of \algname. 
As shown in~\cref{fig:visualization_graph}, the agent of interest has 6 predicted future trajectories for different potential intentions. The dynamic scene graph for each trajectory automatically links the agent to other traffic agents nearby. With these explicit geometry priors, the agent can focus on the interactions with the important agents. From the cases in~\cref{fig:visualization_cases}, \algname enables the ego vehicle to maneuver safely in complex situations like road junction and opposite meeting. These planning abilities results from the accurate motion prediction and necessary inter-agent interactions, based on the proposed graph designs.

\section{Conclusion}
\label{sec:conclusion}

In this paper, we propose a new end-to-end autonomous driving algorithm, GraphAD, which employs an elaborately designed graph to describe heterogeneous interactions in complex traffic scenes. The graph explicitly encodes key driving elements and their relations, allowing us to introduce strong prior knowledge into the algorithm. As a consequence, GraphAD achieves state-of-the-art performance in both the prediction and the planning tasks. 
The way using graphs to encode more complex interactions among diverse traffic instances, such as traffic lights and routing decisions, needs further exploration.

%
%
\bibliographystyle{splncs04}
\bibliography{arxiv}
\end{document}